# Enhancing Confidence Estimation in Telco LLMs via Twin-Pass CoT-Ensembling

Anton Saenko, Pranshav Gajjar, Abiodun Ganiyu, and Vijay K. Shah

*NextG Wireless Lab*, North Carolina State University, Raleigh, NC, USA

{asaenko, prgajjar, aganiyu, vijay.shah}@ncsu.edu

*Abstract*—**Large Language Models (LLMs) are increasingly applied to complex telecommunications tasks, including 3GPP specification analysis and O-RAN network troubleshooting. However, a critical limitation remains: LLM-generated confidence scores are often biased and unreliable, frequently exhibiting systematic overconfidence. This lack of trustworthy self-assessment makes it difficult to verify model outputs and safely rely on them in practice. In this paper, we study confidence calibration in telecom-domain LLMs using the representative Gemma-3 model family (4B, 12B, and 27B parameters), evaluated on TeleQnA, ORANBench, and srsRANBench. We show that single-pass, verbalized confidence estimates fail to reflect true correctness, often assigning high confidence to incorrect predictions. To address this, we propose a novel *Twin-Pass Chain of Thought (CoT)-Ensembling* methodology for improving confidence estimation by leveraging multiple independent reasoning evaluations and aggregating their assessments into a calibrated confidence score. Our approach reduces Expected Calibration Error (ECE) by up to 88% across benchmarks, significantly improving the reliability of model self-assessment. These results highlight the limitations of current confidence estimation practices and demonstrate a practical path toward more trustworthy evaluation of LLM outputs in telecommunications.**

*Index Terms*—**Large language models, Confidence estimation, Telecom, Open Telco LLMs**

## I. INTRODUCTION

The telecommunications industry is undergoing a transformative shift as foundational models, mainly, Large Language Models (LLMs), are increasingly integrated into network operations and management. Modern cellular networks spanning 5G, Open RAN (O-RAN) [1], and emerging 6G architectures are characterized by unprecedented complexity: dense heterogeneous deployments, intricate protocol stacks defined across thousands of 3GPP and other standards specification pages, and the need for rapid adaptation to dynamic traffic patterns and service requirements [2], [3]. This complexity has motivated the adoption of LLMs as cognitive tools capable of reasoning over technical documentation, assisting with troubleshooting, and even automating operational decisions.

Recent work demonstrates LLM applications across a broad spectrum of telecom tasks: automated question answering and specification retrieval from 3GPP and O-RAN documents [4], [5], [6], network troubleshooting and root cause analysis [7], [8], [9], [10], configuration generation and validation [11], [12], [13], intent-based network management [14], and code comprehension for open-source implementations such as srsRAN [7]. These LLM-powered frameworks promise to reduce operational expenditure and accelerate technology deployment by augmenting or even automating tasks traditionally requiring deep domain expertise [2], [3].

An important requirement for deploying LLMs in operational network environments is *reliable confidence estimation*. Not all LLM-assisted tasks require explicit confidence scores; simple information retrieval or document summarization may proceed without them. However, for decision-critical applications, such as autonomously pushing network configurations, triggering mobility parameter adjustments, validating protocol compliance, or executing control actions in programmable RAN architectures, operators must answer a fundamental question: *When a model reports high confidence in its output, can an engineer trust that answer enough to act on it without human verification?* Current deployments typically address this by setting confidence thresholds below which outputs are escalated for human review [15]. The reliability of such thresholds depends entirely on how well the model's reported confidence reflects its actual likelihood of being correct.

A model is considered *well-calibrated* if its predicted confidence exhibits a linear relationship with empirical accuracy: when reporting 80% confidence, it should be correct approximately 80% of the time [16], [17]. Unfortunately, modern LLMs frequently deviate from this ideal, exhibiting systematic overconfidence [18]. In telecommunications, this miscalibration poses acute risks. The dense concentration of specialized terminology, acronyms, and standardized phrasing in 3GPP, O-RAN, ETSI, and ITU documents exacerbates the problem: models trained on such corpora often inflate their certainty simply because generated text mimics the authoritative tone of standards documents, regardless of factual correctness. An overconfident model might autonomously push an incorrect configuration to production, validate a non-compliant protocol implementation, or misdiagnose a network fault, each with potentially severe operational consequences.

While confidence estimation and calibration methods are well-studied in general natural language processing [19], [20], they remain *largely unvalidated* against the unique demands of telecommunications. Existing telecom-LLM benchmarks evaluate accuracy but rarely examine whether reported confidence scores are trustworthy. This gap leaves practitioners without principled guidance for setting operational thresholds, forcing reliance on intuition rather than empirical validation, a recipe for brittle systems that fail unpredictably in production [19]. This paper bridges that gap through a systematic study of LLM confidence calibration on telecommunications-specific bench-marks. We expose the inherent flaws in baseline confidence estimation for technical domains and propose a training-free self-evaluation ensemble method that significantly improves calibration. Our core contributions are:

- We evaluate confidence calibration across three industry-standard benchmarks from the GSMA evaluation suite: Tele-QnA (general telecom), ORANBench (O-RAN specifications), and srsRANBench (5G codebase).
- We propose a training-free method, termed *Twin-Pass CoT-Ensemble*, where the model critiques its own reasoning through multiple stochastic samples, aggregating self-assessed scores to produce calibrated confidence estimates.
- Evaluated with Gemma-3 [21] models across three telecom sub-benchmarks, our proposed methodology achieves up to **88.4%** relative ECE reduction, transforming unreliable confidence scores into actionable operational metrics.
- We provide empirically validated confidence thresholds and show that mean aggregation outperforms median in **55%** of experimental conditions, offering concrete recommendations for telecom practitioners.

## II. BACKGROUND AND RELATED WORK

Large Language Models (LLMs) are autoregressive sequence models that generate text one token at a time. A token can be a full word, a subword unit, or even punctuation, depending on the tokenizer. Given a context sequence $X = (x_1, x_2, \ldots, x_t)$, the model predicts the next token distribution $P(x_{t+1} \mid x_1, \ldots, x_t) = \text{softmax}(z_{t+1})$, where $z_{t+1}$ are the output logits. Decoding repeats this process until a stopping criterion is reached (e.g., end-of-sequence token or length limit). In practice, the prompt determines the task framing and strongly influences both reasoning style and final answer quality [22]. For complex tasks, Chain-of-Thought (CoT) prompting asks the model to produce intermediate reasoning steps before the final answer, often improving multi-step problem solving by making latent reasoning more explicit[23].

Large Language Models operate on the core principle of autoregressive next-token prediction, where a probability distribution is computed over a vocabulary to select the subsequent token. The confidence of an LLM in its generation is traditionally derived from these internal states, specifically through the softmax probability of the predicted tokens, commonly referred to as logprobs [19]. Prior calibration literature shows that token-level probabilities can capture local likelihood more than end-task correctness, especially in complex reasoning settings [19]. In technical domains such as telecommunications, this motivates caution when interpreting high logprob as evidence of factual correctness. To move beyond token-level statistics, research has shifted toward *verbalized confidence*, where the model is prompted to explicitly report its certainty after generating an answer. While this method attempts to capture the model's meta-cognitive assessment, it suffers from severe *generation-induced confirmation bias*. Because the confidence score is generated in the same inference pass as the answer, the model tends to justify its own reasoning trajectory rather than critically evaluating it, leading to systemic overconfidence [18].

Recent studies have shown that while LLMs struggle with calibration in a single pass, they exhibit a significant *evaluation-generation gap*: models are often better at critiquing a reasoning chain than they are at generating the correct answer initially [24], [25]. This suggests that the highest performance and calibration can be achieved by decoupling the response generation from the evaluation process. In the following section, we build upon this insight to propose a method that leverages this gap, demonstrating that multi-pass evaluation of a single reasoning chain provides the most reliable confidence signal for mission-critical telecom operations.

## III. PROPOSED METHODOLOGY

In this section, we introduce the *Twin-Pass CoT-Ensemble*, a methodology designed to break the confirmation bias inherent in single-pass confidence estimation. The name *Twin Pass* refers to the split between the initial generation pass (Pass 1) and the subsequent meta-cognitive evaluation pass (Pass 2). The algorithm 1 formalizes our proposed methodology, which operates in two distinct phases:

**Pass 1: Initial Generation.** An LLM model $M$ receives a telecommunications query $Q$ and generates an answer $A$ along with its step-by-step reasoning $R$. This reasoning identifies the logic used to arrive at the conclusion.

**Pass 2: Blind Meta-Cognitive Evaluation.** In this phase, the same model $M$ is presented with the original question $Q$, the previously generated answer $A$, and the reasoning $R$. Critically, the model is tasked with evaluating the logical soundness of the reasoning $R$ *without* performing a new search or seeing the ground truth. This *blind* critique forces the model to act as an external auditor. To further refine this signal and eliminate stochastic noise from a single evaluation, we perform $N$ independent evaluation passes. In each pass $i$, the model assigns a score $s_i$ based on a detailed QA prompt (Figure 1).

**Aggregation.** The final confidence score, termed the *Ensemble Mean* ($E$), is calculated as the average of these $N$ independent scores: $E = \frac{1}{N}\sum_{i=1}^{N} s_i$. Aggregating multiple independent evaluations of the same reasoning chain reduces stochastic noise and yields a more reliable, better-calibrated confidence signal for telecom engineers.

---

**Algorithm 1** Twin-Pass Chain-of-Thought (CoT) Ensemble

**Require:** Telecom Question $Q$, Language Model $M$, Number of evaluation passes $N$
**Ensure:** Generated Answer $A$, Ensemble Mean $E$, Operational Route
1: **Pass 1: Initial Generation**
2: $A, R \leftarrow$ Generate($M, Q$) {Model $M$ generates Answer $A$ and Reasoning $R$}
3: **Pass 2: Blind Meta-Cognitive Evaluation**
4: $S \leftarrow \emptyset$ {Initialize empty list for scores}
5: **for** $i = 1$ **to** $N$ **do**
6:     $s_i \leftarrow$ Evaluate($M, Q, A, R$) {Blindly rate reasoning from 1 to 5}
7:     $S \leftarrow S \cup \{s_i\}$
8: **end for**
9: **return** $A, E$, Route

```
Meta-Cognitive Evaluation Prompt

You are a Quality Assurance AI for Telecom and O-RAN.

Review the following Question, Answer, and Reasoning.

Question: {question}
Generated Answer: {gen_answer}
Generated Reasoning: {gen_reasoning}

Task:
1. \textbf{Identify Assumptions}: List 1-2 key
   technical assumptions in the reasoning.
2. \textbf{Verify Logic}: Do these assumptions
   logically lead to the answer?
3. \textbf{Rate Confidence}: On a scale of 1-5
   (5=Certain, 1=Unknown).

Return JSON only:
{{
    "assumption_check": "string",
    "self_eval_score": <int 1-5>,
    "self_explanation": "<critique>"
}}
"""
```

Fig. 1. The system prompt utilized during Pass 2 to facilitate the blind meta-cognitive self-evaluation and extract the confidence score ($S_i$).

## IV. EXPERIMENTAL SETUP

### A. Benchmark and Task Setup

To validate our approach, we leverage two standardized open-source telecom benchmarks: **OT-Lite** and **OT-Full**[1], both sourced from the GSMA Open Telecom Benchmarking Suite [26]. Each benchmark contains multiple-choice questions drawn from three complementary sub-benchmarks:

- **TeleQnA** [5]: Questions grounded in 3GPP standards, research publications, and telecom lexicons.
- **ORANBench** [27]: Questions derived from O-RAN Alliance specifications.
- **srsRANBench** [7]: Questions targeting open-source RAN software documentation.

We further select the Gemma-3 model family [21] as our primary LLM $M$. We specifically required an open-weight reasoning LLM to ensure reproducibility while testing our approach across multiple parameter sizes (4B, 12B, and 27B) to verify that our findings are scale-agnostic. Furthermore, Gemma-3 features a substantial 128k-token context window, which provides the necessary capacity to ingest our extended multi-turn reasoning and critique traces without truncation. To maintain computational feasibility the Gemma-3-4B and Gemma-3-27B are evaluated on OT-Lite, while Gemma-3-12B is evaluated on OT-Full.

### B. Evaluation Metric

To quantify calibration quality, we use **Expected Calibration Error (ECE)**, which measures the discrepancy between predicted confidence and empirical accuracy. Lower ECE indicates better alignment between confidence and correctness.

In addition, we analyze the **correlation** between confidence scores and actual correctness measured through accuracy

---

[1]OT-Lite is a curated subset designed for fast iteration, while OT-Full provides the complete question pool for exhaustive evaluation.

across benchmarks to assess whether higher confidence consistently reflects higher accuracy.

### C. Baseline

As highlighted in Section 2, the most prominent method for confidence estimation is a **Single-Pass Verbalized confidence estimation**. We adopt this approach using the prompt template illustrated in Figure 2, which tasks the model with generating both its answer and a corresponding confidence score on a scale of 1 to 5 within a single inference pass.

```
Baseline Prompt

You are a telecommunications expert.
Read the question and options below.

Question: {row['question']}

Options:
{options_text}

Task:
1. Provide a detailed reasoning for the correct answer.
2. Select the correct option (e.g., "option 1").
3. Categorize this question into one of the following
   categories:

{json.dumps(CATEGORIES)}.
4. Rate your confidence in this answer on a scale of
   1-5 (5=Certain, 1=Unknown).

Respond VALID JSON ONLY with this schema:
{{
    "reasoning": "string",
    "answer": "string (e.g. 'option 1')",
    "category": "string",
    "confidence": <int 1-5>
}}
```

Fig. 2. The prompt template utilized by the baseline single-pass verbalized confidence estimation method.

## V. RESULTS

This section presents a systematic analysis of confidence calibration in telecom-domain LLMs. We first demonstrate the failure of single-pass verbalized confidence, then evaluate our proposed Twin-Pass CoT-Ensemble across three benchmarks and three model scales.

### A. The Failure of Single-Pass Verbalized Confidence

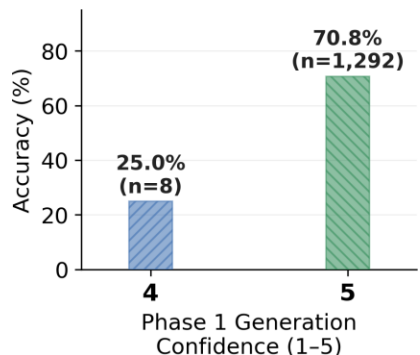

Fig. 3. Accuracy vs. single-pass verbalized confidence (Gemma-3-4B, OT-Lite).

Our analysis revealed that prompting the model to provide a confidence score within the same inference pass is highly ineffective for telecom domain tasks. The verbalized confidence scores were largely indistinguishable between correct and incorrect answers (Figure 3).

The score distributions suffered from severe clustering: essentially all answers, regardless of actual correctness, were rated in the top confidence range which is 4-5. Consequently, there was no meaningful signal to differentiate a correct telecom network configuration from a confident hallucination. This confirms that single-pass confidence reports are consistently high and unreliable. The model tends to assign a high score simply because its response is the most plausible or logically best statement it can generate, rather than because it reflects factual correctness. As a result, the confidence evaluation *is high* even when it is guessing[18].

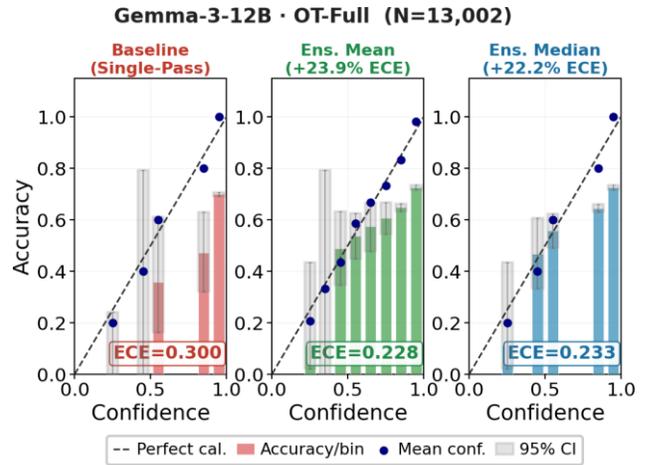

Fig. 5. Reliability diagrams for Gemma-3-12B (OT-Full, $N = 13,002$, all benchmarks pooled). **Left (red):** Raw single-pass verbalized confidence. **Center (green):** Twin-Pass CoT Ensemble Mean ($E$). **Right (blue):** Ensemble Median.

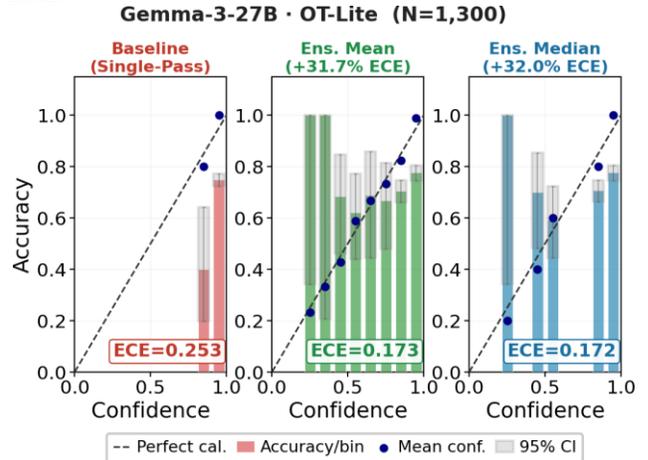

Fig. 6. Reliability diagrams for Gemma-3-27B (OT-Lite, $N = 1,300$, all benchmarks pooled). **Left (red):** Raw single-pass verbalized confidence. **Center (green):** Twin-Pass CoT Ensemble Mean ($E$). **Right (blue):** Ensemble Median.

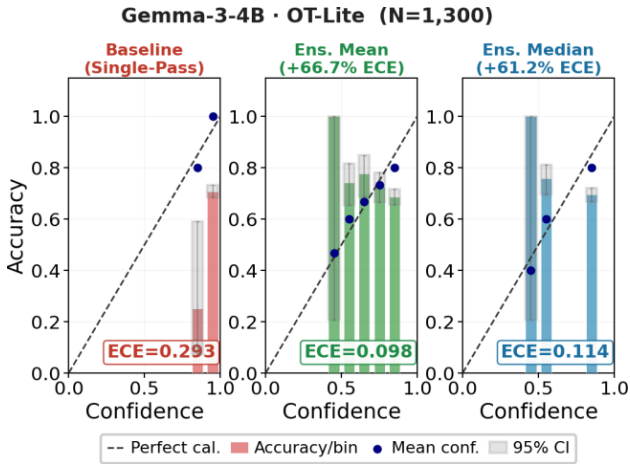

Fig. 4. Reliability diagrams for Gemma-3-4B (OT-Lite, N = 1,300, all benchmarks pooled). **Left (red):** Raw single-pass verbalized confidence. **Center (green):** Twin-Pass CoT Ensemble Mean (E). **Right (blue):** Ensemble Median.

### B. Success of Twin-Pass CoT-Ensemble

Conversely, the Blind Self-Evaluation approach via Twin-Pass CoT-Ensemble demonstrated exceptional discriminative power. By forcing the model to explicitly rate its reasoning in a separate, blind pass and doing so multiple times, we broke the confirmation bias. This achieved a highly calibrated, near-linear separation of predictions, which we term the *Platinum Curve.*

Figure 4 (Gemma-3-4B) provides the first evidence of this transition: compared with the raw single-pass baseline, the ensemble-based curves move substantially closer to the diagonal, indicating improved alignment between confidence and empirical accuracy. The same trend persists at larger model scales in Figure 5 (Gemma-3-12B) and Figure 6 (Gemma-3-27B), where the Twin-Pass ensemble distributions produce more monotonic and interpretable reliability behavior than the baseline confidence signal.

Taken together, Figures 4–6 show a consistent pattern across model sizes and evaluation sets: single-pass verbalized confidence is over-concentrated and weakly discriminative, whereas Twin-Pass aggregation yields confidence tiers that track correctness more faithfully.

Across all evaluated models and datasets, the Twin-Pass CoT-Ensemble produced distinct, actionable reliability tiers revealing consistent scaling behavior across model sizes. Detailed per-benchmark evaluation is reported in Tables I–III. These results form the basis for constructing conservative deployment bounds detailed in Section V-C.

TABLE I
CONFIDENCE RELIABILITY ON TELEQNA. † OT-LITE; ‡ OT-FULL.

| Model | N | Acc. | $ECE_{raw}$ | $ECE_{mean}$ | $ECE_{med}$ | $\Delta ECE$ (%) |
|---|---|---|---|---|---|---|
| Gemma-3-4B† | 1,000 | 0.685 | 0.400 | 0.108 | 0.123 | +73.0 |
| Gemma-3-12B‡ | 10,000 | 0.685 | 0.314 | 0.244 | 0.249 | +22.2 |
| Gemma-3-27B† | 1,000 | 0.743 | 0.255 | 0.172 | 0.172 | +32.5 |

$\Delta ECE$: relative ECE improvement using mean aggregation.

TABLE II
CONFIDENCE RELIABILITY ON **ORANBENCH**. † OT-LITE; ‡ OT-FULL.

| Model | N | Acc. | ECE$_{raw}$ | ECE$_{mean}$ | ECE$_{med}$ | ΔECE (%) |
|---|---|---|---|---|---|---|
| Gemma-3-4B† | 150 | 0.693 | 0.800 | 0.093 | 0.085 | +88.4 |
| Gemma-3-12B† | 1,500 | 0.677 | 0.318 | 0.249 | 0.254 | +21.6 |
| Gemma-3-27B† | 150 | 0.747 | 0.252 | 0.208 | 0.199 | +17.3 |

TABLE III
CONFIDENCE RELIABILITY ON **SRSRANBENCH**. † OT-LITE; ‡ OT-FULL.

| Model | N | Acc. | ECE$_{raw}$ | ECE$_{mean}$ | ECE$_{med}$ | ΔECE (%) |
|---|---|---|---|---|---|---|
| Gemma-3-4B† | 150 | 0.853 | 0.147 | 0.108 | 0.097 | +26.5 |
| Gemma-3-12B† | 1,502 | 0.804 | 0.191 | 0.107 | 0.113 | +43.9 |
| Gemma-3-27B† | 150 | 0.753 | 0.243 | 0.168 | 0.156 | +31.0 |

### C. Golden Threshold Map

Based on these empirical findings, we construct a Golden Threshold Map (Table IV) for deploying the Gemma-3 model family in telecommunications settings.

TABLE IV
GOLDEN THRESHOLD MAP FOR CoT-ENSEMBLE MEAN (EMPIRICALLY VALIDATED)

| Ensemble Mean | Operational Zone | 12B Acc. | 27B Acc. |
|---|---|---|---|
| **4.5 – 5.0** | **Platinum (Auto-Accept)** | 0.73 | 0.78 |
| **4.0 – 4.5** | **Moderate (Consider RAG)** | 0.65 | 0.71 |
| 3.0 – 4.0 | Low Confidence (Route to RAG) | 0.59 | 0.66 |
| < 3.0 | Hallucination Risk (Discard/Flag) | ≤0.44 | ≤0.60 |

This table confirms that for Gemma-3-12B and 27B, the Platinum Tier (ensemble mean ≥4.5) provides a reliable auto-accept criterion with 73–78% empirical accuracy, meaningfully above baseline for both models. Notably, the Gemma-3-4B model never achieves ensemble consensus ≥4.5, reflecting its limited metacognitive capacity at 4B parameters.

## VI. DISCUSSION

The stark contrast between the failure of baseline confidence estimation metrics and the success of metacognitive self-evaluation methodology carries significant implications for RAG pipeline designers in the telecommunications industry. The clustering of logprobs suggests that standard next token prediction metrics are artificially inflated by the dense, standardized jargon inherent to 3GPP and O-RAN documents. Furthermore, the extreme confirmation bias observed in Single Pass Verbalized Confidence demonstrates that LLMs cannot objectively evaluate their own outputs while actively generating them. The model essentially *convinces itself* of its own hallucination. By implementing the proposed *Twin Pass Chain of Thought (CoT) Ensemble approach*, we break this autoregressive bias. Pipeline designers can use this Metacognitive Consensus to implement a Smart Handoff strategy.

Mean aggregation is more effective because it provides greater sensitivity to the scoring signal. While the median collapses outputs to a few integer values, the mean produces a finer-grained scale such as 3.33, 3.67, and 4.25. This improves separation of borderline cases and enables more precise thresholding, while the median remains useful in noisier settings due to its robustness to outliers. Answers scoring in the Platinum Tier (4.5 – 5.0) can bypass human review. Answers in the moderate tier should trigger a secondary RAG loop, prompting the system to retrieve additional contextual chunks from the specifications to reinforce the model's reasoning. Answers with a score below 3 should be immediately escalated to a human engineer or discarded.

An intriguing scaling property emerges when decoupling the authoring and critiquing models. In cross-model evaluation (e.g., forcing a 4B model to critique 12B answers), the smaller model achieves a highly accurate Platinum-tier (∼73%) that it cannot achieve when evaluating its own generations. We attribute this to two factors: (1) verification is a simpler text analysis task than generation, and (2) decoupling breaks the autoregressive inertia that causes smaller models to defend their own mistakes. This demonstrates the economic promise of cascaded systems where small models function as cheap, fast safety gates for high-parameter corporate benchmarks.

## VII. CONCLUSION

This paper presented a large-scale confidence calibration study for Telco LLMs, evaluating the Gemma-3 model family (4B, 12B, and 27B) across the TeleQnA, ORANBench, and srsRANBench sub-benchmarks via the GSMA OT-Lite and OT-Full datasets. We demonstrated that single-pass verbalized confidence metrics provide no reliable signal for identifying hallucinations in domain-specific tasks. Gemma-3-4B exhibited raw ECE values as high as 0.80 on ORANBench, confirming extreme overconfidence. In response, we developed the Twin-Pass CoT-Ensemble architecture, which reduced ECE by over 73% on TeleQnA, 88% on ORANBench, and 26.5% on srsRANBench for Gemma-3-4B, and isolates a *Platinum Tier* of high-reliability answers. Our resulting Golden Threshold Map provides actionable routing logic for network engineers. Future work will focus on integrating this threshold logic directly into active RAG retrieval loops and applying the ensemble methodology to real-time network anomaly detection. We also intend to work towards other model families and parameter scales to see how our approach performs.

## 1. TeleQnA

**Assessment:** Incorrect **Question ID:** teleqna_8938
**Question:** What does CSAT stand for?

---

**Model Generation:**
**Answer:** Option 3
**Reasoning:** CSAT stands for "Coexistence Support for Access Technologies" in cellular networks, detailing operational coexistence between LTE, WiMAX, and NB-IoT to optimize resource utilization.
**Single-Pass Confidence Score:** 5 (Peak confidence).

---

**Critique Ensemble Feedback:**
**Score:** 1.66 ()
**Critique Note:** "The provided reasoning is inaccurate and misleading as CSAT stands for Customer Satisfaction. There is no technical concept in cellular networks containing this abbreviation. The reasoning is completely irrelevant."

Fig. 7. Example of the Critique pipeline identifying and penalizing an acronym hallucination in TeleQnA despite maximum single-pass confidence.

## 2. ORANBench

**Assessment:** Correct **Question ID:** oranbench_1486
**Question:** Which of the following is a measurement type used to track the volume of UL PDCP SDUs received via the Xn-U UL GTP-U tunnel for data forwarding?

---

**Model Generation:**
**Answer:** Option 1
**Reasoning:** Identifies the component keywords required: UL (uplink), Rx (received), Xn-U, and Data Forwarding. Evaluates Option 1 (OR.Xn.UlPdcpSduVolRxXnUUlDataForward) directly mapping to all terms correctly, whilst discounting the DL option (Option 2) and generic extensions (Options 3 and 4).
**Single-Pass Confidence Score:** 5

---

**Critique Ensemble Feedback:**
**Score:** 5.0 ()
**Critique Note:** "The reasoning is highly accurate and logically sound. It correctly identifies the key components within the question and uses these components to evaluate each option. The justification clearly explains why other options aren't correct."

Fig. 8. Example of unanimous meta-cognitive consensus verifying exact component mapping in O-RAN documents.

## 3. srsRANBench

**Assessment:** Incorrect **Question ID:** srsranbench_203
**Question:** What is the number of bits to modulate in the PBCH modulator?

---

**Model Generation:**
**Answer:** Option 3
**Reasoning:** Identifies QPSK means 2 bits per symbol. Iteratively recalculates 16 * 2 = 32 bits iteratively in a circle, and then concludes: "None of the options match this value... so Option 3 (512) seems closest to a reasonable value although it isn't correct."
**Single-Pass Confidence Score:** 3

---

**Critique Ensemble Feedback:**
**Score:** 2.0 ()
**Critique Note:** "The reasoning goes astray very quickly. It gets stuck in loops recalculating bits repeatedly, despite acknowledging none of them match. The jump to Option 3 as a 'closest reasonable value' is unsupported and indicates confusion rather than logical deduction."

Fig. 9. Example of the ensemble detecting mathematical looping and unsupported guessing logic within the reasoning chain.

This appendix provides detailed qualitative examples of the Twin-Pass CoT-Ensemble in action across the three evaluated sub-benchmarks: TeleQnA, ORANBench, and srsRANBench. These examples illustrate how the blind critique passes can identify and penalize confident hallucinations, as well as correctly validate rigid reasoning chains.

As illustrated in Figure 7, this example showcases the Critique pipeline successfully rescuing the evaluation metric from a plausible-sounding hallucination. The model initially overclaimed a highly descriptive technical definition for a non-existent acronym, a mistake that could easily slip through standard automatic string parsers. However, the ensemble accurately detects the fabrication and correctly lowers the evaluation to a failure state.

Figure 8 demonstrates a scenario where a unanimous 5.0 meta-cognitive agreement is achieved. The critique explains that the model did not merely guess the correct answer, but instead rigidly filtered out opposing options based on exact acronym compliance, such as correctly distinguishing between downlink (DL) and uplink (UL) interfaces.

Finally, Figure 9 highlights the ensemble's capacity for reasoning-level debugging. Rather than relying on simple keyword lookups, the Critique pass identifies an initially accurate setup that quickly devolves into arithmetic looping and unsupported guessing. By recognizing exactly where the model's logic breaks down and becomes confused, the ensemble correctly penalizes the confidence score.